\documentclass{llncs}

\usepackage{graphicx}
\usepackage{subfigure} 
\usepackage{color}
\usepackage{amsmath}
\usepackage{amssymb}
\usepackage{algorithm}
\usepackage{algorithmic}
\usepackage{wrapfig}
\usepackage{txfonts}
\usepackage{mathdots}
\usepackage{wrapfig}

\DeclareMathAlphabet{\mathcal}{OMS}{cmsy}{m}{n}
\newcommand{\R}{\mathbb{R}}
\newcommand{\Argmin}[1]{\underset{#1}{\textrm{arg~min }}}
\newcommand{\PiS}[1]{\mathcal{P}_\mathcal{S} \left(#1\right)}
\renewcommand{\matrix}[1]{\textbf{#1}}
\renewcommand{\vector}[1]{\textbf{#1}}

\newcommand{\citep}{\cite}
\newcommand{\citet}{\cite}

\DeclareMathOperator{\Gr}{Gr}
\DeclareMathOperator{\St}{St}

\definecolor{gray}{rgb}{.5,.5,.5}

\begin{document}
\titlerunning{Robust SLRA on the Grassmannian}  % abbreviated title (for running head)

\author{Clemens Hage \and Martin Kleinsteuber}
\authorrunning{C. Hage and M. Kleinsteuber} % abbreviated author list (for running head)
\institute{Department of Electrical Engineering and Information Technology\\
Technische Universit\"{a}t M\"{u}nchen\\
Arcisstr.~21, 80333 Munich, Germany \\
\email{\{hage,kleinsteuber\}@tum.de},\\
\texttt{http://www.gol.ei.tum.de}}

\title{Robust Structured Low-Rank Approximation on the Grassmannian}

\maketitle              % typeset the title of the contribution

\begin{abstract}
Over the past years Robust PCA has been established as a standard tool for reliable low-rank approximation of matrices in the presence of outliers. Recently, the Robust PCA approach via nuclear norm minimization has been extended to matrices with linear structures which appear in applications such as system identification and data series analysis. At the same time it has been shown how to control the rank of a structured approximation via matrix factorization approaches. The drawbacks of these methods either lie in the lack of robustness against outliers or in their static nature of repeated batch-processing. We present a Robust Structured Low-Rank Approximation method on the Grassmannian that on the one hand allows for fast re-initialization in an online setting due to subspace identification with manifolds, and that is robust against outliers due to a smooth approximation of the $\ell_p$-norm cost function on the other hand. The method is evaluated in online time series forecasting tasks on simulated and real-world data.
\end{abstract}

\section{Introduction}
Many applications such as system identification and time series analysis motivate the problem of Structured Low-Rank Approximation (SLRA). While common low-rank approximations like PCA aim to find a low-dimensional subspace to represent high-dimensional data optimally with respect to some norm or divergence, in the structured case this problem is extended by the additional constraint that the low-rank approximation has to meet a certain linear structure (Hankel, Toeplitz, Sylvester).

For the prominent case of Hankel matrices a method dubbed Singular Spectrum Analysis (SSA) \citep{Broomhead1986} has been presented, which performs the simplest way of SLRA in that it computes a low-rank approximation of a (Hankel-) structured matrix followed by a so-called diagonal averaging step, which is the projection onto the space of Hankel matrices. An obvious drawback of this method is that this projection can destroy the low-rank property established before. The method of Cadzow \citet{Cadzow1988} alternates between these two steps until the algorithm converges to a solution that is indeed low-rank and structured. However, as Chu et al.~\citet{Chu2003} and Markovsky~\citet{Markovsky2008} state, this solution can be far away from the initialization with no guarantees of finding an actually meaningful approximation to the data.
Recently, Ishteva et al.~\citet{Ishteva2014} have proposed a factorization approach with a cost function that joints the structural and low-rank constraint. The dimension of the two matrix factors is an upper bound on the rank of the approximation and its structure is enforced with a side condition. The approximation is fitted to the data according to an $\ell_2$-norm, although it is well known that low-rank approximations of this kind can be vulnerable against outliers. For the unstructured case this has been the major incentive to move from PCA to robustified PCA methods such as the Robust PCA method by Candes et al.~\citet{Candes2011} which recovers a subspace in the presence of sparse outliers of great magnitude. Ayazoglu et al.~\citet{Ayazoglu2012} have proposed to extend this concept to structured matrices by introducing additional Lagrangian multipliers. Their method is called Structured Robust PCA (SRPCA), and its performance is demonstrated in visual applications like Target Location Prediction, Tracklet Matching (both matrix completion problems) and Outlier Removal from trajectories, which can be interpreted as outlier identification through robust subspace estimation.

One of the drawbacks of many Robust PCA approaches and thus also the SRPCA approach is their batch-processing nature. In an online setting the algorithm needs to be re-run from scratch if the data set grows or changes over time. This can be alleviated by factorizing the low-rank approximation of data into an orthogonal matrix representing the subspace and a coefficient matrix containing the coordinates of the currently observed data in this subspace, cf.~\citep{Balzano2010,Hage2014} for the unstructured case. Whenever new data comes in, it is possible to initialize the subspace optimization with the previous estimate and to update both the subspace and the coordinates to the new data. Obviously, whenever the subspace does not change significantly this saves computational effort compared to a random initialization.
We will firstly derive a batch algorithm for Robust Structured Low-Rank Approximation on the Grassmannian and then outline how to process structured data online in an efficient way. We illustrate the performance of the proposed algorithm on several time series forecasting tasks.

\section{Robust Structured Low-Rank Approximation on the Grassmannian using a smoothed $\ell_p$-norm cost function}
\subsection{Low-rank and sparsity constraints in the unstructured case}

Low-rank approximation of data is a well-studied problem, cf.~\cite{Markovsky2014} for a recent overview. In the past years a a trend can be seen towards Robust PCA methods that are tolerant against outliers in the data. An often considered data model is $\matrix{X} = \matrix{L} + \matrix{S}$, where the input data $\matrix{X} \in \mathbb{R}^{m \times n}$ is assumed to be composed of a low-rank part $\matrix{L}$ with $\textrm{rank}\left(\matrix{L}\right)\leq k$ and a sparse matrix $\matrix{S}$ that contains few non-zero entries. While in \citet{Candes2011} the low-rank constraint is enforced via minimization of the nuclear norm, in this work we will consider the Robust PCA setting on the \emph{Grassmannian}
\begin{align}
\Gr_{k,m} := \{\matrix{P} \in \mathbb{R}^{m \times m} | \matrix{P} = \matrix{U} \matrix{U}^\top, \matrix{U} \in \St_{k,m}\}.
\end{align}
The subspace is hereby represented by an element $\matrix{U}$ of the \emph{Stiefel} manifold $\St_{k,m} = \{\matrix{U} \in \mathbb{R}^{m \times k} | \matrix{U}^\top \matrix{U} = \matrix{I}_k\}$, where $\matrix{I}_k$ is the $(k \times k)$-identity matrix. The low-rank approximation is then $\matrix{L} = \matrix{U}\matrix{Y}$ with $\matrix{Y} \in \mathbb{R}^{k \times n}$.

In contrast to the $\ell_2$-norm in common PCA, robust approaches use relaxations of the $\ell_0$-norm for fitting the low-rank approximation to the data, as gross outliers might otherwise distort the estimation. The smoothed $\ell_p$-norm
\begin{align}
\label{eq:lpnorm}
h_{\mu} \colon \R^{m \times n} &\to \R^+, \quad X \mapsto \sum_{j=1}^n\sum_{i=1}^m \left(x_{ij}^2 + \mu\right)^\frac{p}{2}, \quad 0 < p < 1
\end{align}
presented in \cite{Hage2014} behaves similarly to the $\ell_0$-norm for sparse outliers, but treats small additive Gaussian noise like an $\ell_2$ norm. A Robust Low-Rank Approximation problem using the sparsifying function \eqref{eq:lpnorm} writes as
\begin{align}
\label{eq:rpca_model}
\left(\hat{\matrix{U}},\hat{\matrix{Y}}\right) = \arg \min_{\matrix{U} \in \St_{m,k}, \matrix{Y} \in \mathbb{R}^{k \times n}} h_\mu (\matrix{X} - \matrix{U}\matrix{Y}), \quad \quad \hat{\matrix{L}} = \hat{\matrix{U}}\hat{\matrix{Y}}.
\end{align}
The sparse component can be recovered via $\hat{\matrix{S}} = \matrix{X}-\hat{\matrix{L}}$, possibly followed by a thresholding operation to remove residual noise.
\subsection{Extension to linear matrix structures and algorithmic description}
\label{sec:SLRAG}
Besides the low-rank and sparsity decomposition no other assumptions are made on the data in \eqref{eq:rpca_model}. In many applications however, structured matrices like Hankel, Sylvester or Toeplitz matrices play an important role. Following the notation of \citet{Ishteva2014} we denote by $\mathcal{S}$ a linear matrix structure, by $\mathcal{S}(\vector{d})$ a structured matrix obtained from a data series $\vector{d}$ and by $\PiS{\matrix{X}}$ the orthogonal projection of any (possibly unstructured) matrix $\matrix{X}$ onto the image of $\mathcal{S}$ w.r.t.~the standard inner product. For example, if $\mathcal{H}$ is a Hankel structure the orthogonal projection $\mathcal{P}_\mathcal{H}$ is equivalent to the diagonal averaging step in SSA \citep{Broomhead1986}.

In order to include the structural constraint we extend the cost function \eqref{eq:rpca_model} by the side condition $\matrix{L} \in \mathcal{S}$. This motivates the Lagrangian Multiplier scheme
\begin{align}
\min_{\matrix{U} \in \St_{m,k}, \matrix{Y} \in \mathbb{R}^{k \times n}, \Lambda \in \mathbb{R}^{m \times n}} h_\mu \left(\matrix{X} - \matrix{UY}\right) + \langle \Lambda , \matrix{UY} - \PiS{\matrix{UY}} \rangle + \tfrac{\rho}{2} \| \matrix{UY} - \PiS{\matrix{UY}} \|_F^2.
\end{align}

Algorithm \ref{alg:batchcase} outlines the extension of the Robust PCA method from Hage and Kleinsteuber \cite{Hage2014} to the case of structured matrices. The algorithm considers a partial observation $\hat{\matrix{X}}$ of the data with $\mathcal{A}$ defining which entries are actually observed. In each iteration, three steps are performed. Firstly, the subspace estimate is updated. In this realization the subspace is uniquely identified with a Grassmannian projector $\matrix{P} = \matrix{U}\matrix{U}^\top$ with $\matrix{U} \in \St_{k,m}$. The optimization problem is solved via Conjugate Gradient (CG) descent with backtracking line-search using a QR-decomposition based retraction on the Grassmannian. Once $\matrix{P}$ and thereby $\matrix{U}$ have been found the coordinates $\matrix{Y}$ are updated with a CG method in Euclidean space, such that a new optimum low-rank estimate $\matrix{L} = \matrix{U}\matrix{Y}$ is found. In a third step the Lagrangian multiplier $\Lambda$ is updated, then $\rho$ is increased and $\mu$ is decreased. While $\mu$ controls the behavior of the sparsifying function $h_\mu (\cdot)$, the parameter $\rho$ weighs between the data fitting term and the structural side condition. More precisely, as long as $\rho$ is small the Robust PCA term is the leading power and a low-rank approximation is fitted to the data. With increasing $\rho$ the structural condition is more and more enforced until it is the dominating term in the cost function.
\begin{wrapfigure}{r}{0.5\textwidth}
\begin{minipage}{0.5\textwidth}
\vspace{-0.75cm}
\begin{algorithm}[H]
   \caption{Alternating minimization scheme for Grassmannian Robust Structured Low-Rank Approximation}
\label{alg:batchcase}
\begin{algorithmic}
\vspace{6pt}
   \STATE {\bfseries Initialize:}
   \STATE Choose $\matrix{X}_0 \in \mathbb{R}^{m \times n}$, s.t.~$\mathcal{A}(\matrix{X}_0) = \hat{\matrix{X}}$.
   \STATE Initialize $\matrix{U}^{(0)}$ randomly or from $k$ left singular vectors of $\matrix{X}_0$.
   \vspace{6pt}
   \STATE $\matrix{Y}^{(0)} = \matrix{U}^{(0)\,\top} \matrix{X}_0$, $\matrix{L}^{(0)} = \matrix{U}^{(0)}\matrix{Y}^{(0)}$
   \STATE $\matrix{P}^{(0)} = \matrix{U}^{(0)}\matrix{U}^{(0)\,\top}$
   \vspace{6pt}
   \STATE Choose $\mu^{(0)}$ and $\mu^{(I)}$, compute $c_\mu = \left(\tfrac{\mu^{(I)}}{\mu^{(0)}}\right)^{1/(I-1)}$
   \STATE Choose $\rho^{(0)}$ and $\rho^{(I)}$, compute $c_\rho = \left(\tfrac{\rho^{(I)}}{\rho^{(0)}}\right)^{1/(I-1)}$
   \vspace{6pt}
\FOR{$i = 1 : I$}
\vspace{6pt}
   \STATE $\matrix{P}^{(i+1)} = \Argmin {\matrix{P} \in \Gr_{k,m}} h_{\mu^{(i)}}(\hat{\matrix{X}} - \mathcal{A}(\matrix{P}\matrix{L}^{(i)})) \newline\phantom{\matrix{P}^{(i+1)} = }- \left\langle \Lambda^{(i)},\matrix{P} \matrix{L}^{(i)} - \PiS{\matrix{P} \matrix{L}^{(i)}} \right\rangle \newline\phantom{\matrix{P}^{(i+1)} = }+ \tfrac{\rho^{(i)}}{2} \|\matrix{P} \matrix{L}^{(i)} - \PiS{\matrix{P} \matrix{L}^{(i)}} \|_F^2$
   \vspace{6pt}
   \STATE find $\matrix{U}^{(i+1)}$ \; s.t. \; $\matrix{U}^{(i+1)}\matrix{U}^{(i+1)\,\top} = \matrix{P}^{(i+1)}$
   \begin{flushright}\textbf{Subspace Step}\end{flushright}
   \vspace{-6pt}
\textcolor{gray}{\hrule}
\vspace{3pt}
   \STATE $\matrix{Y}^{(i+1)} = \Argmin {\matrix{Y} \in \R^{k \times n}} h_{\mu^{(i)}}(\hat{\matrix{X}} - \mathcal{A}(\matrix{U}^{(i+1)} \matrix{Y})) \newline\phantom{\matrix{Y}^{(i+1)} = }- \left\langle \Lambda^{(i)}, \matrix{U}^{(i+1)} \matrix{Y} - \PiS{\matrix{U}^{(i+1)} \matrix{Y}} \right\rangle \newline\phantom{\matrix{Y}^{(i+1)} = }+ \tfrac{\rho^{(i)}}{2} \|\matrix{U}^{(i+1)} \matrix{Y} - \PiS{\matrix{U}^{(i+1)} \matrix{Y}} \|_F^2$
\vspace{6pt}
\STATE $\matrix{L}^{(i+1)} = \matrix{U}^{(i+1)}\matrix{Y}^{(i+1)}$ \begin{flushright}
\vspace{-6pt}
\textbf{Coordinate Step} \end{flushright}
\vspace{-6pt}
\textcolor{gray}{\hrule}
\vspace{6pt}
\STATE $\Lambda^{(i+1)} = \Lambda^{(i)} - \rho^{(i)} \left( \matrix{L}^{(i+1)} - \PiS{\matrix{L}^{(i+1)}} \right)$ \begin{flushright}
\textbf{Multiplier Update}
\end{flushright}
\vspace{-6pt}
\textcolor{gray}{\hrule}
\vspace{3pt}
   \STATE $\mu^{(i+1)} = c_\mu \mu^{(i)}, \quad \rho^{(i+1)} = c_\rho \rho^{(i)}$
   \vspace{6pt}
   \STATE $\epsilon = \tfrac{1}{mn} \|\matrix{L}^{(i+1)} - \PiS{\matrix{L}^{(i+1)}} \ \|_F$
\vspace{6pt}
\ENDFOR
\vspace{6pt}
\STATE $\hat{\matrix{L}} = \PiS{\matrix{L}^{(I)}}, \quad \hat{\matrix{S}} = \matrix{X} - \hat{\matrix{L}}$
\end{algorithmic}
\end{algorithm}
\vspace{-3cm}
\end{minipage}
\end{wrapfigure}
In a practical application the optimization can also be terminated if the residual Hankel penalty $\epsilon$, i.e.~the Frobenius distance to the next Hankel-structured matrix normalized by the number of matrix entries falls below a certain threshold $\tau$.

\section{Efficient Online Time Series Forecasting via Robust SLRA}
We have outlined how to use manifold optimization for Robust SLRA on the Grassmannian. In the important case of a Hankel structure SLRA corresponds to identifying an LTI system, cf.~\cite{Markovsky2014}. In practical applications, however, an observed system might be time-variant. Or the observed data is not related to a physical system at all but still exhibits repetitive or periodic behavior. The field of Time Series Analysis \citep{Box1976} deals with these signals and numerous auto-regressive methods for filtering and forecasting data series exist. In the SLRA context a low-rank Hankel matrix (and thus an LTI) is fitted to the observed data and the future development is extrapolated from this approximation. Thereby, the rank bounds the complexity of the approximation. If the behavior of the data changes over time a new model needs to be determined for each observation instance. For our proposed Grassmannian Robust SLRA method this means that both a new subspace and new coordinates need to be computed. However, when the signal characteristics vary moderately over time it is likely that the new subspace lies close to the previously found one. Therefore, the subspace should not be randomly initialized but rather updated with the new data point, in a similar way as Robust Subspace Tracking (\citep{He2012}, \citep{Seidel2014}) in the unstructured case. As discussed earlier, however, we do not optimize directly on the space of structured low-rank matrices. Instead we relax the structural constraint, update the subspace and then tighten the structural side condition again by varying the parameter $\rho$ in the cost function. 
\begin{wrapfigure}{r}{0.5\textwidth}
\begin{minipage}{0.5\textwidth}
\vspace{-0.75cm}
\begin{algorithm}[H]
   \caption{Online Time Series Forecasting via Grassmannian Robust SLRA}
\label{alg:forecasting}
\begin{algorithmic}
\STATE \textbf{Input:} data series $\vector{d} \in \mathbb{R}^N$, system order $k$, analysis dimension $m$, forecasting range $r$
\FOR{$j = 2m:N$}
\vspace{6pt}
   \STATE Define $\vector{x}_{(j)} = \begin{bmatrix}\vector{d}(j-(2m-1) \colon j)^\top & | &  \vector{0}_r^\top\end{bmatrix}^\top$ and $\mathcal{A}$ according to forecasting range
   \vspace{6pt}
   \STATE Obtain $\matrix{X}_{(j)} = \mathcal{H}(\vector{x}_{(j)})$
   \STATE Initialize $\matrix{U}^{(0)}=\matrix{U}_{(j-1)}$   
   \STATE $\matrix{Y}^{(0)} = \matrix{U}^{(0)\,\top} \matrix{X}_{(j)}$, $\matrix{L}^{(0)} = \matrix{U}^{(0)}\matrix{Y}^{(0)}$
   \STATE $\matrix{P}^{(0)} = \matrix{U}^{(0)}\matrix{U}^{(0)\,\top}$
\STATE Select number of iterations $I_{(j)}$
   \STATE Choose $\rho^{(0)}$ and $\rho^{(I)}$
   \STATE Compute $c_{\rho\,(j)} = \left(\tfrac{\rho^{(I)}}{\rho^{(0)}}\right)^{1/(I_{(j)}-1)}$
\FOR{$i = 1:I_{(j)}$}
\vspace{6pt}
\STATE \textbf{Subspace Step} from Alg.~\ref{alg:batchcase}
\STATE \textbf{Coordinate Step} from Alg.~\ref{alg:batchcase}
\STATE \textbf{Multiplier Update} from Alg.~\ref{alg:batchcase}
\STATE $\rho^{(i+1)} = c_{\rho\,(j)} \rho^{(i)}$
\ENDFOR
\vspace{6pt}
\STATE $\vector{l}_{(j)} = \PiS{\matrix{L}^{(I_{(j)})}}$
\STATE $\hat{\vector{d}}_{(j)}\left(j+1 \colon j+r\right) = \vector{l}_{(j)}\left(2m \colon 2m+r-1\right)$
\vspace{6pt}
\ENDFOR
\end{algorithmic}
\end{algorithm}
\vspace{-1.5cm}
\end{minipage}
\end{wrapfigure}
%\begin{bmatrix} x_1^{(0)} & x_1^{(1)}& \cdots & x_1^{(n-1)} & \vrule & \textcolor{red}{x_1^{(n)}} \\
%x_2^{(0)} & x_2^{(1)}& \cdots & x_2^{(n-1)} & \vrule & \textcolor{red}{x_2^{(n)}} \\
%\vdots & \vdots & \ddots & \vdots & \vrule & \textcolor{red}{\vdots} \\
%x_m^{(0)} & x_m^{(1)}& \cdots & x_m^{(n-1)} & \vrule & \textcolor{red}{x_1^{(n)}} \end{bmatrix} \quad \quad \quad

In Algorithm~\ref{alg:forecasting} we describe an Online Time Series Forecasting method based on Robust Structured Low-Rank Approximation on the Grassmannian. The algorithm receives as inputs the time series of data $\vector{d}$ to be analyzed as well as the desired order of the system and the forecasting range, i.e.~the number of samples to be predicted. Since a Hankel matrix of size $m \times m$ contains $(2m-1)$ samples of data, the prediction starts at $\vector{d}(2m)$. A data vector $\vector{x}_{(j)}$ of length $(2m-1)$ is extracted from the data up to the present position, padded with zeros according to the forecasting range and structured to form a Hankel matrix. The first subspace estimate $\matrix{U}^{(0)}$ is not initialized randomly but with $\matrix{U}_{(j-1)}$, the final subspace estimate of the previous set of data samples. Note that the subscript \emph{(j)} counts the position of the current set of data samples in the data stream while in Algorithm~\ref{alg:batchcase} the superscript \emph{(i)} denoted the iteration count of the alternating minimization steps in the batch process. Accordingly, for each set of data samples at position $j$ the number of alternating minimization steps $I_{(j)}$ for the current estimation must be set beforehand. This number varies between a predefined $I_{min}$ and $I_{max}$, and the choice is based on the residual Hankel penalty $\epsilon_{(j-1)}$ of the previous iteration. This corresponds to the observation that significant changes in the system behavior lead to higher values for $\epsilon$ and require more iterations in the optimization process, whereas less update steps are required if the subspace changes slowly or does not change at all. We start with the previous iterate on the Stiefel manifold and execute the three steps from Algorithm~\ref{alg:batchcase} in an alternating manner until convergence. Notice that the cost function parameter $\mu$ is fixed here and only the Lagrangian parameter $\rho$ is changed in each iteration to steer between data fit and structure. The forecasting is realized as a robust matrix completion problem, i.e.~the respective entries in the lower right corner of $\matrix{X}$ are considered unobserved entries. Once a structured low-rank estimate has been found, the predicted entries of $\vector{d}$ can easily be read from the last entries of $\vector{l}_{(j)}$.

\section{Experimental results}
In a first experiment we evaluate our method on an impulse response prediction task for a noisy observation of a simulated SISO Linear Time Varying (LTV) system. The data is generated via
\begin{align*}
\dot{\vector{x}}(t+1) &= \matrix{A}(t)\vector{x}(t) + \vector{b}^\top \vector{u}(t), \quad \matrix{A}(t) = e^{0.001 t\, \matrix{Z}}, \quad \matrix{Z}^\top = -\matrix{Z}\\
\vector{y}(t) &= \vector{c}^\top \vector{x}(t) + \vector{n}(t)
\end{align*}
\begin{figure}[t]
\includegraphics[width=\textwidth]{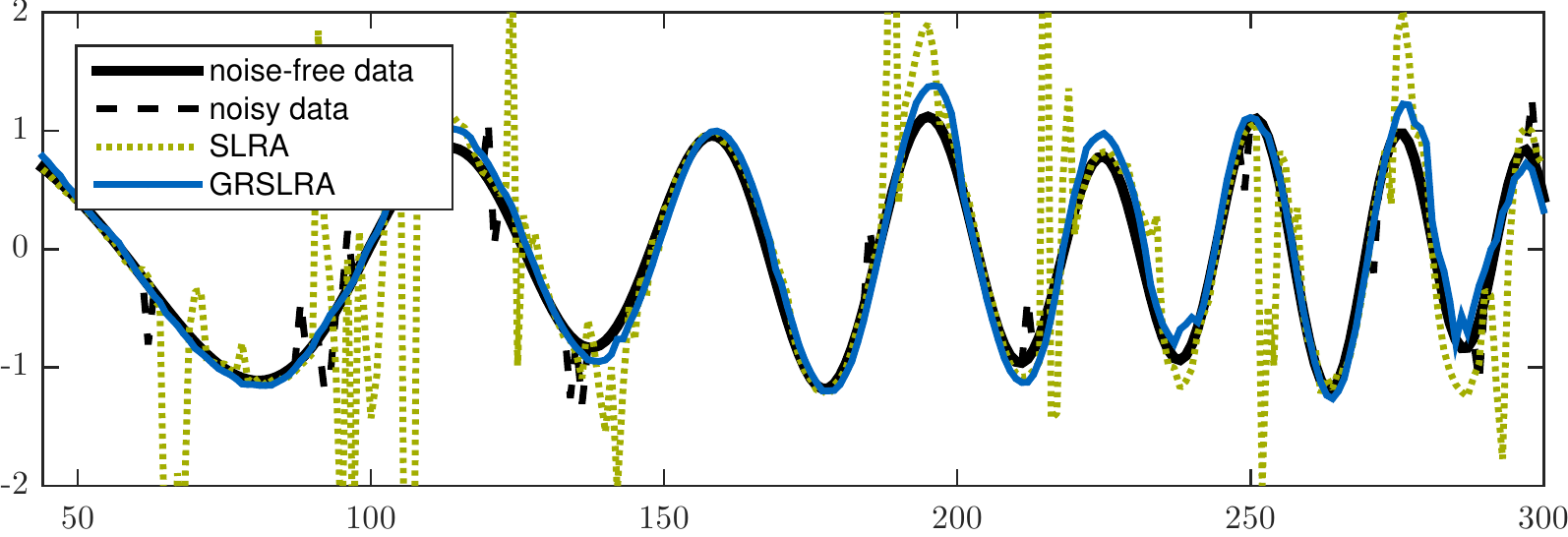}
\caption{Forecasting of 3 samples of a SISO-LTV impulse response with additive noise and ouliers}
\label{fig:ltv}
\end{figure}
with $\vector{b}, \vector{c}$ being uniform random vectors and $\matrix{Z}$ being a random skew-symmetric matrix. The additive observation noise $\vector{n}(t)$ contains of two parts, Gaussian noise with $\sigma=0.01$ and randomly appearing (rate $0.05$) salt and pepper noise samples that take on values of $\pm0.5$. The degree of the system is chosen as $k=5$ and we generate the impulse response of the system for $300$ samples. Our method is compared to the SLRA method from \citet{Ishteva2014} and both algorithms are implemented in {\emph{MATLAB}} on a desktop computer. We predict three time steps into the future from an observation of $2m-1$ samples, and the parameters are empirically chosen as $m=20$, $\rho \in [10^{-6},\ 10]$ (both methods), $p=0.5, \mu=0.005, \tau=5 \times 10^{-4}$. The SLRA method is randomly initialized in each step, converges within $30$~iterations and requires about $0.4$~seconds. The iteration number of our method varies between $I_{min}=16$ and $I_{max}=128$ iterations with an average of $22$~iterations that add up to $0.7$~seconds per forecasting step. The forecasting results in Figure~\ref{fig:ltv} indicate that both methods are able to cope with the Gaussian noise quite well and predict the system behavior quite reliably, but the spurious outliers introduce errors in the SLRA extrapolation due to the $\ell_2$-error weighting. Our proposed method is much more robust at the price of a higher computational effort. However, due to the beneficial subspace initialization the computation time is still competitive.

\begin{figure}[t]
\includegraphics[width=\textwidth]{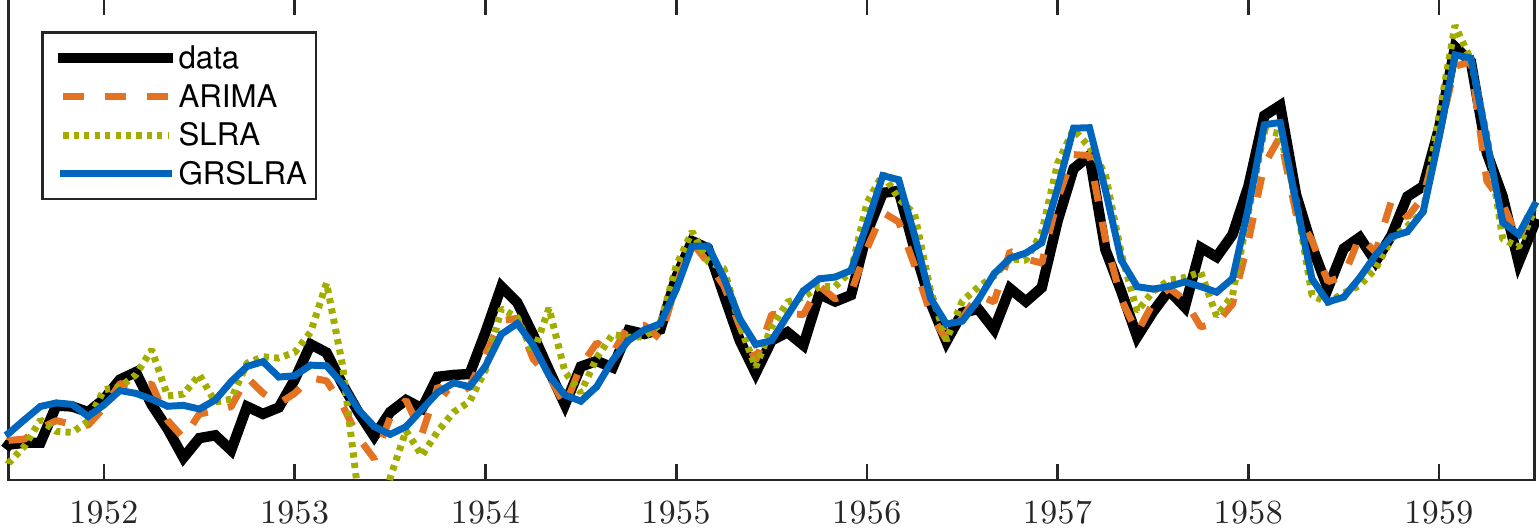}
\caption{Six month forecast of monthly airline passenger data from the years 1952-1960 based on 3 year observation period.}
\label{fig:passengers}
\end{figure}

\begin{figure}[t]
\includegraphics[width=\textwidth]{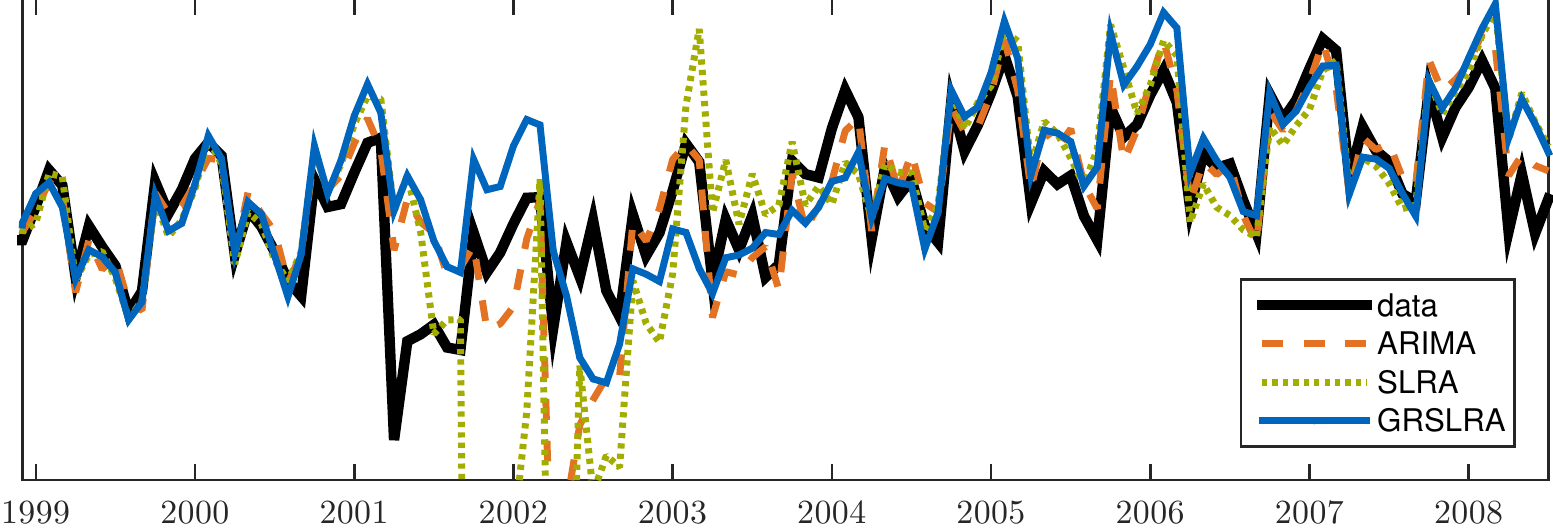}
\caption{Six month forecast of monthly airline passenger data from the years 1996-2009 based on 3 year observation period.}
\label{fig:passengersnew}
\end{figure}

In a second experiment we compare our method on real-world data with SLRA and the \emph{forecast} routine in MATLAB with a 12-month-seasonal ARIMA(0,1,1) model\footnote{\url{http://mathworks.com/help/econ/forecast-airline-passenger-counts.html}}. The time series is the well-known \emph{Airline Passenger} dataset from \cite{Box1976} normalized to the range $[0\; 1]$. The upper bound on the rank of the approximation is chosen as $k=8$, and we forecast $6$ samples from $2m-1$ samples with $m=18$, which corresponds to projecting the monthly amount of passengers half a year into the future from observing the past three years. Figure~\ref{fig:passengers} shows that all three methods succeed in forecasting the data, with average absolute deviations of $0.060$ for SLRA, $0.036$ for ARIMA and $0.044$ for our method. On average, the ARIMA implementation requires $1.3$s, SLRA $0.7$s and our method ($I_{min}=8, I_{max}=64$, $12$ iterations on average) is the fastest with $0.3$s. The popularity of this well-known but also well-behaving dataset has inspired us to perform another experiment on airline passenger data. We have obtained the system-wide (domestic and international) number of passenger emplanements in the USA for the years $1996-2014$ from the American Bureau of Transportation Statistics \cite{BTS2015}. Due to the dramatic developments in the year $2001$ this data is obviously more challenging. Figure~\ref{fig:passengersnew} shows the dataset and the six month forecasts of the three compared methods with the experimental setup as before.
The seasonal ARIMA model copes best with the challenging conditions (average absolute error of $0.086$), but it needs to be noticed that the actual seasonality is known a priori while SLRA and our method do not have this information. As before, the SLRA method is able to forecast data reliably under good conditions but suffers from the gross outliers, resulting in an average absolute error of $0.182$. Finally, our method shows more robustness against gross outliers (average absolute error of $0.102$), although in this real-world example the low-rank and sparse data model is not exactly met.

\section{Conclusion}
We have presented a novel method for Robust Structured Low-Rank Approximation on the Grassmannian. Using an approximated $\ell_p$-norm, the method robustly fits an approximation of upper-bounded rank and linear structure to the given data. For the special case of a Hankel structure we have furthermore shown how to use the developed concept for Robust Online Time Series Forecasting. We have shown how to benefit from the manifold setting in online processing, as we can increase the efficiency by re-using the previously identified subspace. Experimental results show that our method performs effectively and efficiently in simulated and real-world applications.

\bibliography{slra_paper}

\end{document}